\documentclass[runningheads]{llncs}

\usepackage[table,xcdraw]{xcolor}

\usepackage{caption}

\usepackage{cite}
\usepackage{subcaption}
\usepackage{adjustbox}
\usepackage{tikz}
\usepackage{times}
\usepackage{enumitem} 

\usepackage{multirow}

\usepackage[colorlinks=true, allcolors=blue]{hyperref}
\usepackage{graphicx}

\makeatletter

\begin{document}

\mainmatter

\title{AI-Enhanced Business Process Automation:\\ A Case Study in the Insurance Domain Using Object-Centric Process Mining}

\titlerunning{AI-Enhanced Business Process Automation}

\author{
    Shahrzad Khayatbashi\inst{1},
    Viktor Sjölind\inst{2},
    Anders Granåker\inst{2},
    Amin Jalali\inst{2,3}}
\authorrunning{Shahrzad Khayatbashi et al.}  

\institute{
    Linköping University, Linköping, Sweden,
    \email{shahrzad.khayatbashi@liu.se}
        \and 
    If P\&C Insurance,
    \email{viktor.sjolind@if.fi, anders.granaker@if.se, amin.jalali@if.se}
        \and
    Stockholm University, Stockholm, Sweden,
    \email{aj@dsv.su.se}
}

\maketitle

\begin{abstract}   
Recent advancements in Artificial Intelligence (AI), particularly Large Language Models (LLMs), have enhanced organizations' ability to reengineer business processes by automating knowledge-intensive tasks. This automation drives digital transformation, often through gradual transitions that improve process efficiency and effectiveness. To fully assess the impact of such automation, a data-driven analysis approach is needed — one that examines how traditional and AI-enhanced process variants coexist during this transition. Object-Centric Process Mining (OCPM) has emerged as a valuable method that enables such analysis, yet real-world case studies are still needed to demonstrate its applicability. This paper presents a case study from the insurance sector, where an LLM was deployed in production to automate the identification of claim parts, a task previously performed manually and identified as a bottleneck for scalability. To evaluate this transformation, we apply OCPM to assess the impact of AI-driven automation on process scalability. Our findings indicate that while LLMs significantly enhance operational capacity, they also introduce new process dynamics that require further refinement. This study also demonstrates the practical application of OCPM in a real-world setting, highlighting its advantages and limitations.

\keywords {AI-Driven Automation, Business Process Reengineering, Digital Transformation, Business Process Management}
\end{abstract}

\vspace{-1.8\baselineskip}	
    
\section{Introduction} \label{sec:intro}
\vspace{-0.3\baselineskip}	
Artificial Intelligence (AI), particularly in the form of Large Language Models (LLMs)\cite{brown2020language}, is transforming business processes by automating knowledge-intensive tasks and enhancing operational efficiency. For example, identifying claim parts in the insurance domain is a knowledge-intensive task that traditionally requires human expertise due to the wide variety of claim parts. As claim volumes increase, manual identification becomes a bottleneck, limiting process scalability. 
LLMs present a significant opportunity to automate this task by leveraging their ability to understand context, recognize entities, and classify textual information based on prior training on vast datasets. Specifically, LLMs can be fine-tuned or prompted to extract relevant parts and classify them accurately within diverse claim descriptions, reducing reliance on human input and supporting digital transformation in business process management.

Organizations often transition from traditional (as-is) processes to AI-driven (to-be) workflows gradually, resulting in a phase where both process variants operate in parallel. 
This coexistence allows organizations to evaluate the impact of AI automation on their ongoing processes, but it also presents challenges, as both process variants can influence the process outcomes. 
To support this, a data-driven approach should analyze both parallel process variants, each representing distinct aspects of the process, thereby enabling a more comprehensive understanding of how automation alters process behavior~\cite{jalali2014aspect}.
Object-Centric Process Mining (OCPM)~\cite{van2019object} has emerged as a promising method for analyzing process transformations that enables the simultaneous analysis of multiple perspectives, making it particularly valuable for AI-driven process reengineering. 
Investigating side effects of process automation is important as indicated in literature~\cite{vu2023towards,bider2016limiting}, where OCPM enable such investigation based on recorded process data.
While previous research has demonstrated OCPM’s use in analyzing a single business process variant~\cite{berti2023analyzing,park2023analyzing,kretzschmann2024overstock}, its ability to capture both as-is and to-be process variants simultaneously has yet to be investigated.

To address this gap, this study presents a real-world case from the insurance sector, where the rising volume of claims necessitated AI-driven automation for claim part identification. The study follows the Business Process Reengineering (BPR) framework~\cite{bpr1998} to examine the transition from a manual as-is process to an AI-enhanced to-be workflow, supported by an LLM~\cite{brown2020language} for automation. Instead of fully automating claim part management, the LLM model generates leads for investigators, ensuring a balance between AI-driven efficiency and expert oversight and having humans in the loop. The AI implementation adheres to the CRoss-Industry Standard Process for Data Mining (CRISP-DM) methodology~\cite{crisp2000}, providing a structured approach to data understanding, modeling, and deployment.

To evaluate the impact of AI-driven automation, this study applies OCPM to assess process scalability. The evaluation follows OCPM$^2$~\cite{ocpm22025}, an extension of the PM$^2$ process mining methodology~\cite{van2015pm}, enabling a systematic examination of process evolution. By applying OCPM, this study provides empirical insights into how AI-driven process reengineering enhances scalability while also assessing OCPM’s strengths and limitations in analyzing real-world process transformations.

The remainder of this paper is structured as follows: Section~\ref{sec:method} details the methodology, outlining the application of BPR, CRISP-DM, and OCPM for process evaluation. Section~\ref{sec:result} presents the results, examining the impact of AI-driven automation on scalability and process dynamics while evaluating OCPM’s advantages and limitations. Finally, Section~\ref{sec:conclusion} concludes the paper.
\vspace{-0.5\baselineskip}	
\section{Methodology}\label{sec:method}
\vspace{-0.5\baselineskip}	
\begin{figure}[t!]
    \centering
    \adjustbox{center}{\includegraphics[width=1.1\linewidth]{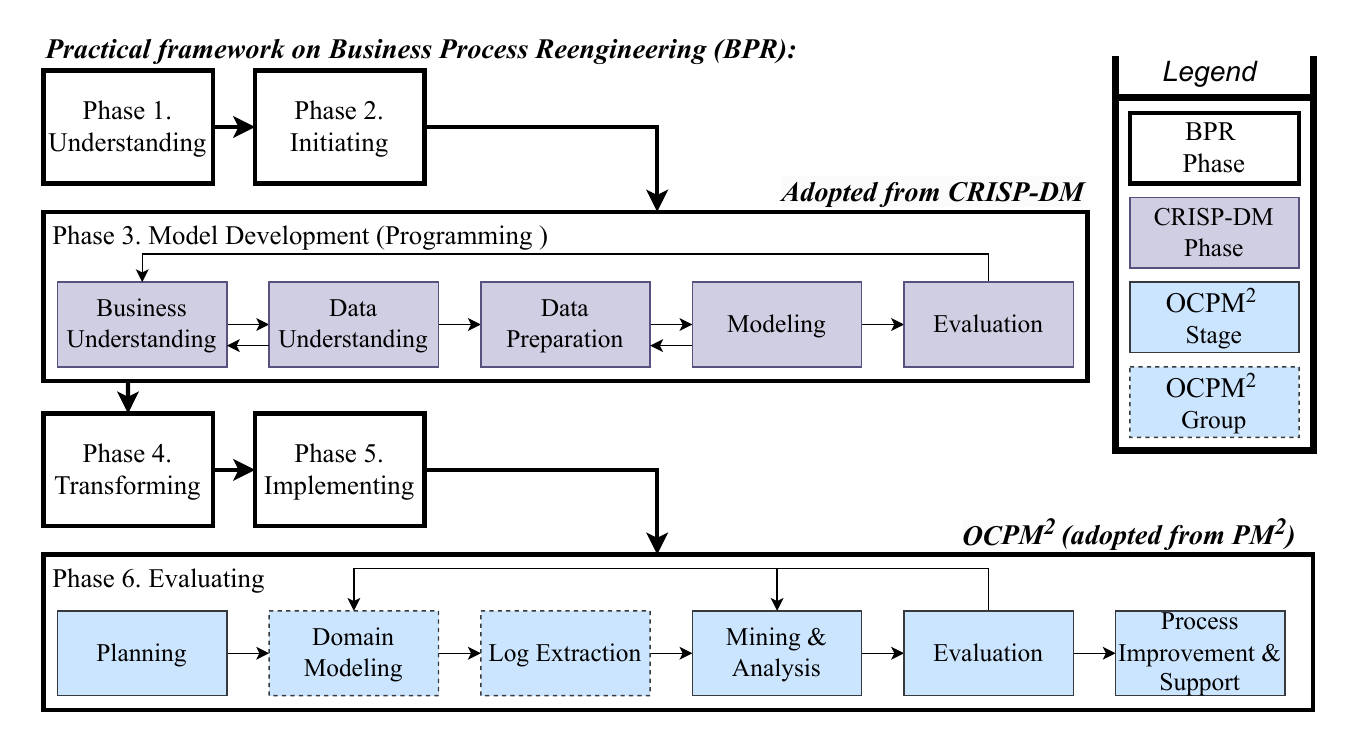}}
    % \vspace{-1.5\baselineskip}	
    \caption{The followed research methodology based on BPR~\cite{bpr1998}, CRISP-DM~\cite{crisp2000} and OCPM$^2$~\cite{ocpm22025}, an extended version of PM$^2$~\cite{van2015pm}.}
    \label{fig:method}
    \vspace{-1.8\baselineskip}	
\end{figure}

\figurename~\ref{fig:method} illustrates the Business Process Reengineering (BPR) framework~\cite{bpr1998} consisting of 6 phases that are followed in this study.
In the \textit{understanding phase}, the objectives of reengineering were defined, emphasizing the role of AI-driven automation in enhancing the scalability of the claim part identification process. This phase involved identifying scalability limitations and key challenges by analyzing the as-is process. Stakeholder interviews and process documentation were conducted to understand manual claim assessments, their associated bottlenecks, and the feasibility of AI-driven automation.

Following this, the \textit{initiation phase} defined the project vision, scope, and requirements. The primary goal was to develop an AI-driven approach for automating the identification of claim parts requiring further investigation. Additionally, data collection strategies were outlined to ensure necessary historical claim data availability for AI model training.

Next, the \textit{model development phase} (referred to as Programming in BPR) was carried out by the technical team, following the CRISP-DM methodology~\cite{crisp2000} to ensure a structured approach in developing a customized AI model, where its tasks are explained below:
\begin{itemize}[leftmargin=*]
    \item \textit{Business Understanding}, where the objectives and constraints of AI integration within the claims handling process were defined. Security approval was obtained to ensure compliance with security and privacy regulations. Discussions with domain experts helped establish requirements for claim part identification, ensuring alignment with operational goals. The primary business objective was to automate knowledge-intensive tasks while maintaining acceptable performance and enhancing scalability.
    \item \textit{Data Understanding}, where data sources, data quality issues, and relevant features were identified. The analysis focused primarily on claim descriptions and notes, representing the unstructured data sources. Additionally, exploratory data analysis (EDA)~\cite{camizuli2018exploratory} was conducted to gain deeper insights.
    \item \textit{Data Preparation}, a labeled dataset was created by asking claim part investigators to label free-text claim descriptions and notes. This unstructured data served as the ground truth for evaluating the quality of the developed AI solution.
    \item \textit{Modeling}, where we leveraged the analytical capabilities of Large Language Models (LLMs) to identify claim parts. Various approaches were evaluated to ensure a consistent output format from GPT models~\cite{achiam2023gpt}, with the Structured Output feature in OpenAI APIs found to be the most effective \cite{openai_docs}. The prompt engineering process was a collaboration between business domain experts and data scientists.
    \item \textit{Evaluation}, where the LLM model was assessed using two approaches. First, quantitative evaluation was performed against the ground truth dataset. Second, a language evaluation was conducted to analyze the impact of using the Finnish language versus English on the model's performance. Each claim was processed in Finnish and English by automatically translating the input and combining it with an English version of the instructions. Business validation was further conducted through pilot tests, where human experts reviewed the AI-generated outputs before full deployment.
\end{itemize}

The \textit{transformation phase} involved conducting pilot studies to evaluate AI integration, estimating the scale of organizational changes, and assessing resource requirements.

The \textit{implementation phase} focused on restructuring workflows and integrating AI into the IT infrastructure. The solution was implemented using serverless components, queuing mechanisms, table storage, and models hosted in an Azure OpenAI Service, ensuring that GPT models operated within a regional data center in Sweden to comply with company policies on data privacy and security. Additional privacy measures were applied, such as masking Personally Identifiable Information (PII) before transmitting data to the AI inference server.

Finally, the \textit{evaluation phase}  aimed to assess the impact of AI on process scalability using process mining techniques, following OCPM$^2$~\cite{ocpm22025}, an extended version of the PM$^2$~\cite{van2015pm} methodology. The process began with the planning stage, where we formulated the key research questions.
Next, in the domain modeling, we designed the domain by identifying object types, activities, and their relationships. Following this, we extracted an Object-Centric Event Log (OCEL  2.0)~\cite{berti2024ocel} and applied object-centric and traditional process discovery techniques to analyze the AI-driven transformation. 
OCEL was selected as the data format due to its support for advanced analytical operations such as drill-down, unfolding~\cite{khayatbashi2024advancing,khayatbashi2024olap}, and ad-hoc filtering (e.g., dicing), which were essential for exploring different process perspectives and ensuring flexible, performance-efficient querying during analysis~\cite{khayatbashi2023transforming}.
The findings were then empirically validated by domain experts to ensure reliability, and we ultimately derived insights to enhance the claims part management process.

\vspace{-0.5\baselineskip}	
\newcommand{\caseTotalCustomers}{4 million}
\newcommand{\caseTotalClaims}{1.4 million}
\newcommand{\caseTotalClaimsPayout}{115 million}
\newcommand{\caseTotalLOB}{77}

% from time stamps
\newcommand{\caseDateProjectStart}{2024-01-30}
\newcommand{\caseDateFirstModelDeveloped}{2024-04-16} 
\newcommand{\caseDateDeployed}{2024-09-25} 
\newcommand{\caseNumberOfModelUpdates}{4} 
\newcommand{\caseDataCollectionPeriodInMonth}{5}

% from dfg
\newcommand{\csDFGrc}{3743}
\newcommand{\csDFGcn}{3743}
\newcommand{\csDFGpCPIncomingFlow}{1034}
\newcommand{\csDFGcCPi}{26}

% humen
\newcommand{\csHDFGrCP}{68}
\newcommand{\csHDFGrCPpercentage}{1.82\%} %
\newcommand{\csHDFGcCPi}{21}

% ai
\newcommand{\csAIDFGpCPIncomingFlow}{1034}
\newcommand{\csAIDFGrCPpercentage}{27.62\%} % 
\newcommand{\csAIDFGcCPi}{23}

\newcommand{\caseAIvsHumanCPIdentifiedDiffernece}{1420\%} % 100 *(csAIDFGpCPIncomingFlow - csHDFGrCP) / csHDFGrCP

\newcommand{\csAIMissed}{3}
\newcommand{\csHMissed}{5}
\newcommand{\csAIandHuman}{18} % = csDFGcCPi - two above

\section{Case Study}\label{sec:result}
\vspace{-0.5\baselineskip}	
\subsection{Process Context}
\vspace{-0.5\baselineskip}	
The case study was conducted at \textit{If P\&C Insurance}, a property and casualty insurance company serving approximately \caseTotalCustomers{} customers across the Nordic and Baltic regions~\cite{if_insurance}. The company offers a comprehensive range of insurance solutions and services tailored to diverse customer segments, including individuals and large corporations. It processes more than \caseTotalClaims{} claims annually, with half of them handled within 24 hours of being reported. This study focuses on a sub-process within the claims handling procedure, specifically aimed at identifying certain types of \textit{claim parts} that require special treatment. Due to confidentiality restrictions, the business terms for these claim parts cannot be disclosed.

This business process operates in the Nordic and Baltic regions across three business areas, each comprising multiple lines of business. In total, there are \caseTotalLOB{} lines of business where different variations of this process are in place. The process heavily relies on claim handlers to identify claim parts. However, the variations across lines of business make it challenging to train claim handlers, as identifying these components requires specialized business knowledge. In this study, we applied AI to identify claim parts in one specific line of business.

The goal to develop an AI-driven solution was set by the business before the start of the project, based on prior internal analysis and strategic digital transformation objectives. However, the BPR and CRISP-DM phases were essential for tailoring the solution, validating feasibility, aligning stakeholder expectations, preparing the data, and evaluating the impact of the AI system in practice.
The project began on \caseDateProjectStart{}, with the first version of the model developed on \caseDateFirstModelDeveloped{}. The model was integrated and deployed in production on \caseDateDeployed{}. This study evaluates the impact of this process reengineering on the scalability of the business process, based on data collected over \caseDataCollectionPeriodInMonth{} months during which both AI and claim handlers worked in parallel to identify claim parts.
Claim handlers continued to work as before, even after the introduction of AI, allowing us to assess its impact on the scalability of the business process. 

This section presents the results from phases 3 and 6 of the Business Process Reengineering (BPR) project. It includes both the evaluation outcomes and the main lessons learned during the study.
A total of twelve people from the company took part in the project, enabling the implementation of phases 1 to 5 and supporting the first author to evaluate the outcome using OCPM. 
The team included two data scientists, one software engineer, two subject matter experts, one business owner, one DevOps specialist, one cloud engineer, one data privacy officer, one legal expert, one identity and access management (IAM) expert, one enterprise architect, and one application security specialist.
The first author led the work in Phase 6, focusing on the process mining analysis, with support from the second and last authors in providing the required data.
The second and third authors were responsible for conducting Phase 3.

\vspace{-0.5\baselineskip}	
\subsection{Phase 3: Model Development}
The development of the claim part identification model followed the CRISP-DM methodology, ensuring a structured and iterative approach. Close collaboration between technical team and domain experts played a crucial role in refining requirements and improving model accuracy. Given the importance of precise claim part identification in automated processing, the goal was to develop a model capable of achieving at least human-level performance. 
The company’s baseline analysis revealed that human claim handlers correctly identify 70\% of the actual claim parts, meaning that 30\% of the true items are missed. This recall value sets a performance benchmark against which the LLM models were evaluated.
A balanced dataset was manually labeled by claim part investigators to ensure that model predictions could be assessed reliably.

To evaluate the LLM models, we used Recall as the primary evaluation metric. It was prioritized over F1-score because, in claim processing, false negatives (missed claim parts) have a significantly higher impact than false positives. While the F1-score balances precision and recall, it does not align with real-world priorities, where failing to detect claim parts has a tangible negative financial impact. In contrast, false positives can often be handled through additional validation steps with minimal disruption. Since the consequences of false negatives outweigh those of false positives, maximizing recall is more critical than achieving a perfect balance. By focusing on Recall, the model ensures a higher likelihood of identifying all relevant claim parts, even if some additional filtering is required downstream, which will be handled by investigators.

In the initial Proof-of-Concept, we tested a variety of different models, including the small language model (SLM) Phi-3 from Microsoft and the large language model (LLM) Mistral Large. Our tests showed that these models did not perform adequately when processing input data and prompts in the Finnish language.
Since the prompts expose business-specific terminology that is subject to confidentiality restrictions, we are unable to disclose them in this paper.
For the production system, we eventually settled on GPT-4o-0806, employing a Chain-of-Thought (CoT) prompt structure with few-shot samples to guide the model in identifying claim parts, and different strategies to produce an output format that could easily be integrated into the system. GPT-4o also seemed to give similar levels of performance in both Finnish and English. Throughout the project, therefore, we kept running two separate analysis processes for each claim task: one in which we translated both the input data and prompts from Finnish to English and another where we ran the analysis using input data and prompts in Finnish.
The input data, mainly claim descriptions and notes, were masked to eliminate Personally Identifiable Information (PII) before processing by the LLM. In the English version, we also used GPT-4o-0806 to translate input data before processing. All LLMs in use were deployed in an Azure OpenAI service with data residency in Sweden.

\begin{table}[t!]
\centering
\begin{tabular}{|c|l|l|l|l|l|}
\hline
\multicolumn{1}{|l|}{Model Version} & Language                             & Accuracy & Precision & Recall                                & F1\_score \\ \hline
                                    & eng                                  & 0.76     & 0.83      & 0.67                                  & 0.74      \\ \cline{2-6} 
\multirow{-2}{*}{v1}                & fin                               & 0.77     & 0.86      & 0.67                                  & 0.75      \\ \hline
                                    & eng                                  & 0.77     & 0.92      & 0.61                                  & 0.73      \\ \cline{2-6} 
\multirow{-2}{*}{v2}                & fin                               & 0.76     & 0.85      & 0.64                                  & 0.73      \\ \hline
                                    & eng                                  & 0.7      & 0.8       & 0.56                                  & 0.66      \\ \cline{2-6} 
\multirow{-2}{*}{v3}                & \cellcolor[HTML]{A0E09F}fin       & 0.77     & 0.81      & \cellcolor[HTML]{A0E09F}\textbf{0.72} & 0.76      \\ \hline
                                    & eng                                  & 0.71     & 0.79      & 0.61                                  & 0.69      \\ \cline{2-6} 
\multirow{-2}{*}{v4}                & \cellcolor[HTML]{A0E09F}fin      & 0.74     & 0.76      & \cellcolor[HTML]{A0E09F}\textbf{0.72} & 0.74      \\ \hline
                                    & \cellcolor[HTML]{A0E09F}eng & 0.8      & 0.81      & \cellcolor[HTML]{A0E09F}\textbf{0.81} & 0.81      \\ \cline{2-6} 
\multirow{-2}{*}{v5}                & fin                               & 0.77     & 0.83      & 0.69                                  & 0.76      \\ \hline
\end{tabular}
    \vspace{1\baselineskip}	
    \caption{Evaluation results of different model versions for Claim Part identification.}
    \label{tbl:ModelEvaluation}
    \vspace{-2.5\baselineskip}	
\end{table}

As the model evolved through multiple iterations, systematic improvements were made to enhance performance. 
\tablename~\ref{tbl:ModelEvaluation} shows the evaluation result for different model versions. As can be seen, version 1 and version 2 fell short of human performance, with recall scores lagging behind the 70\% baseline. 
The initial model used an XML format to produce a stable output format. This approach would occasionally produce slight variations in the output when we expected binary results for a task. In the second version of the model, we switched to the JSON mode API feature~\cite{openai_docs_json}, which guaranteed that the output was valid JSON, and although there were improvements, we still had not fully solved the issue of unwanted variations in the output.

By version 5, instruction sets were significantly enriched with detailed metadata considerations, such as insurance event timing, which played a crucial role in improving claim part likelihood estimation. We also solved the issue of unwanted variations in the output using the OpenAI Structured Outputs feature that forced the output to adhere to a strict schema~\cite{openai_docs}.
A major refinement in version 5 was the introduction of specialized claim part prompts, which allowed the model to assess the existence of specific components separately, leading to improved prediction scores.
The changes enabled the model (version 5) to achieve an 81\% recall for the English version, surpassing human performance and demonstrating its ability to automate claim processing with a high degree of accuracy. This milestone was particularly critical for the business, as it allowed for large-scale automation while maintaining consistency in claim assessments.

\vspace{-0.5\baselineskip}	
\subsection{Phase 6: Evaluation}
\vspace{-0.5\baselineskip}	
We followed OCPM$^2$~\cite{ocpm22025}, an extended version of the PM$^2$~\cite{van2015pm} methodology, to evaluate the impact of our AI-driven business process reengineering.
\vspace{-0.5\baselineskip}	
\subsubsection{Planning:}
In this stage, we aimed to answer the following four questions:
\begin{itemize}[leftmargin=*]
    \item \refstepcounter{enumi} \label{q:1} \textit{Q1) How have claim handlers effectively identified claim parts?}
    \item \refstepcounter{enumi} \label{q:2} \textit{Q2) How has AI effectively identified claim parts?}
    \item \refstepcounter{enumi} \label{q:3} \textit{Q3) Among the claim parts under investigation, how many claim parts did AI fail to identify but were accurately identified by claim handlers?}
    \item \refstepcounter{enumi} \label{q:4} \textit{Q4) Among the claim parts under investigation, how many claim parts did claim handlers fail to identify but were accurately identified by AI?}
\end{itemize}
Additionally, we identified various systems from which we needed to extract data.

\vspace{-0.5\baselineskip}	
\subsubsection{Domain Modeling:}
\begin{figure}[t!]
    \centering
    \includegraphics[width=1\linewidth]{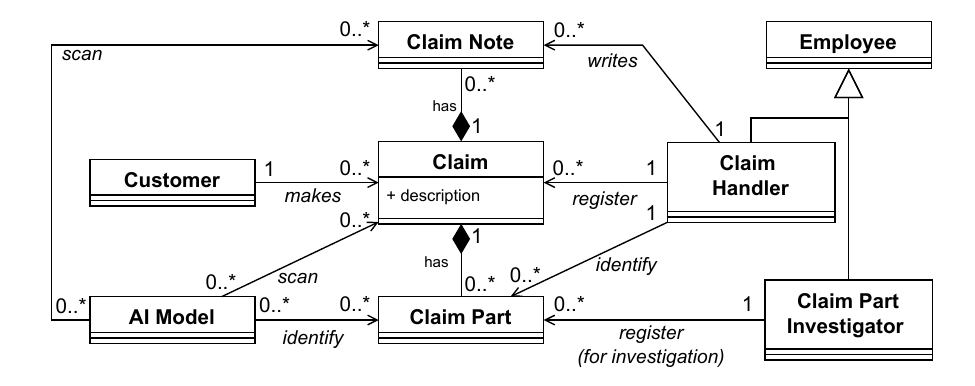}
    \caption{The conceptual model illustrating the identified object types and their relationships within the claims part identification process.}
    \label{fig:ConceptualModel}
    \vspace{-1.5\baselineskip}	
\end{figure}

Through multiple iterations, we identified relevant object types based on the questions and designed a business conceptual model in UML notation, illustrating the object types and their relationships within this business process context (see \figurename~\ref{fig:ConceptualModel}). The identified object types are \textit{Customer}, \textit{Claim}, \textit{Claim Note}, \textit{Claim Part}, \textit{AI Model}, and \textit{Employee}, where an employee can be either a \textit{Claim Handler} or a \textit{Claim Part Investigator}.
All these object types are defined as classes, with \textit{Claim Handler} and \textit{Claim Part Investigator} inheriting from the \textit{Employee} class.

A customer can make multiple claims, but each claim can be made by only one customer. Each claim can be registered by a claim handler, while a claim handler can register multiple claims. Each claim can have multiple claim notes, and each note is written by a claim handler. A claim handler can also identify claim parts, where each claim part belongs to a specific claim. A claim can have multiple claim parts.
An AI model can scan claim descriptions and claim notes to identify claim parts. Additionally, a claim part investigator can register a claim part for further investigation.

In addition, we documented the relation between activities and object types using an ``Extraction Matrix", where activities were listed as rows and object types as columns. We do not present this matrix in this paper, as we will show the existence of relations later when we elaborate on verifying the extraction. 

In this paper, we used the following acronyms to refer to activities and some object types for the sake of the readability of figures, i.e.: \textit{\textbf{CP}: Claim Part}, \textit{\textbf{rc}: register claim}, \textit{\textbf{cn}: create note}, \textit{\textbf{rCP}: report Claim Part}, \textit{\textbf{cCPi}: create Claim Part investigation}. These two acronyms refer to activities that enhance the process using AI, i.e., \textit{\textbf{sc}: scan claim}, \textit{\textbf{pCP}: predict Claim Part}.
In the traditional process, after registering the claim and creating notes, a claim handler could identify and report claim parts. Then, a claim part investigator could create a claim part investigation for them, which initiates another process. 
The integration of AI introduced two additional activities: the AI model scans the claim (reading claim descriptions and claim notes), and it identifies the claim part (referred to as the ``predict claim part").
Then, the claim part investigator can register a claim part investigation for them as well. 

\subsubsection{Implementation and Log Extraction:}
Extracting logs posed several challenges despite the presence of multiple data warehouses within the company. Fortunately, the company maintained a temporal data warehouse that stored not only claim-related data but also tracked changes to each entity. However, information about individual claim parts had to be extracted from an operational system using a specific KPI.
We extracted temporal data that captured all relevant information but filtered the results to retain only one note per claim—the most recent note available prior to the scan claim activity (\textit{sc}). This filtering step facilitated the transformation of data into the OCEL 2.0 format, as this standard cannot distinguish whether related claim notes have expired.
This limitation could be addressed by defining a validity period for object-to-object relations within OCEL 2.0. Notably, this limitation does not exist in all OCED formats, e.g., temporal Event Knowledge Graphs (tEKG)~\cite{khayatbashi2024transforming} capture snapshots of objects and their relationships at a particular point in time.

\begin{figure}[t!]
    \centering
    \includegraphics[width=0.7\linewidth]{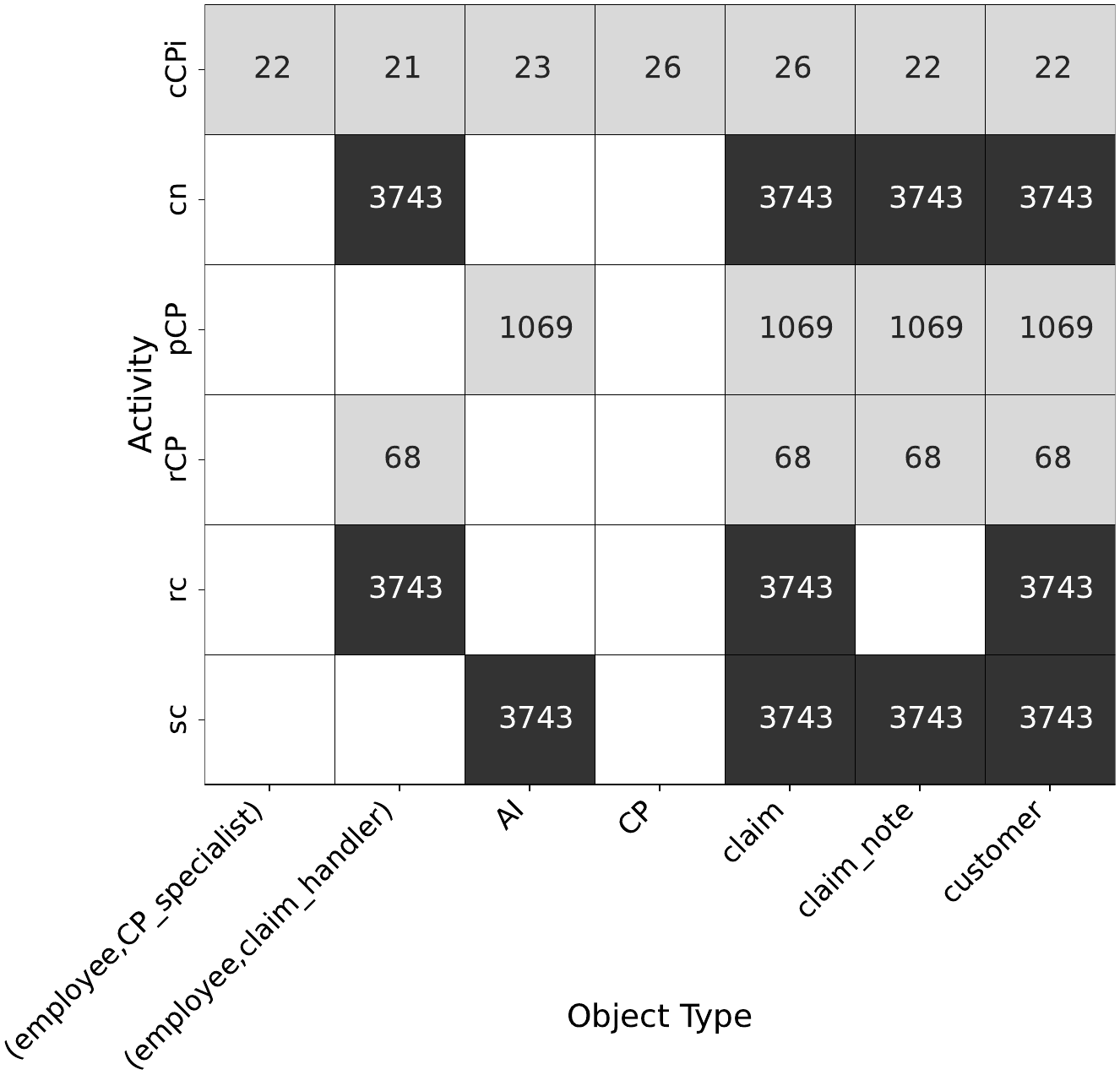}
    \vspace{-.7\baselineskip}	
    \caption{The verification matrix showing the number of relations between activities and object types in the extracted OCEL with these acronyms used for activity names: \textit{\textbf{rc}: register claim}, \textit{\textbf{cn}: create note}, \textit{\textbf{rCP}: report Claim Part}, \textit{\textbf{cCPi}: create Claim Part investigation}, \textit{\textbf{sc}: scan claim}, \textit{\textbf{pCP}: predict Claim Part}. \textit{\textbf{CP}} is also used as an acronym for \textit{Claim Part} object type.}
    \label{fig:heatmap}
    \vspace{-1.5\baselineskip}	
\end{figure}

We performed a drill-down operation at the employee role level to distinguish between claim handlers and claim part investigators aiming to discover more detailed patterns~\cite{2025uncovering_patterns} when analyzing the process. We also verified the log by checking whether all relations identified in the Extraction Matrix were captured. \figurename~\ref{fig:heatmap} shows the number of relations captured in extracted OCEL 2.0 between each activity and different object types, depicted in the corresponding cells. This heatmap was used to verify the correctness of the extracted OCEL 2.0.

\subsubsection{Process Mining Analysis and Evaluation:}

We analyzed the results by discovering the Object-Centric Directly-Follows Graph (OC-DFG)~\cite{van2019object} and Object-Centric Petri Nets (OCPNs)~\cite{van2020discovering} from the extracted OCEL. We found OC-DFG particularly useful for analysis, as we needed to examine all possible paths that could be directly followed to perform different analyses, which we elaborated on later.
However, \textit{using any object-centric process model to communicate with stakeholders proved ineffective}. Stakeholders found these models too complex to interpret, even when we reduced the number of object types by selecting a more specific profile (i.e., a set of object types). Despite our efforts to simplify the visualizations, stakeholders still found them difficult to interpret during the final presentation.
As a result, while we used OC-DFG to analyze different scenarios, we flattened the filtered OCELs by focusing on the Claim object type (which was related to all activities) and presented the results using a Directly-Follows Graph (DFG). 
Flattening has been considered as an effective technique to complement OCPM~\cite{jalali2022object}.
In this study, stakeholders found these graphs informative, as they improved communication when discussing and investigating different scenarios.
Although object-centric models enabled deeper analysis and log filtering for specific insights, DFGs proved more effective for conveying results to end users.

\begin{figure}[t!]
    \centering
    \adjustbox{center}{\includegraphics[width=1.2\linewidth]{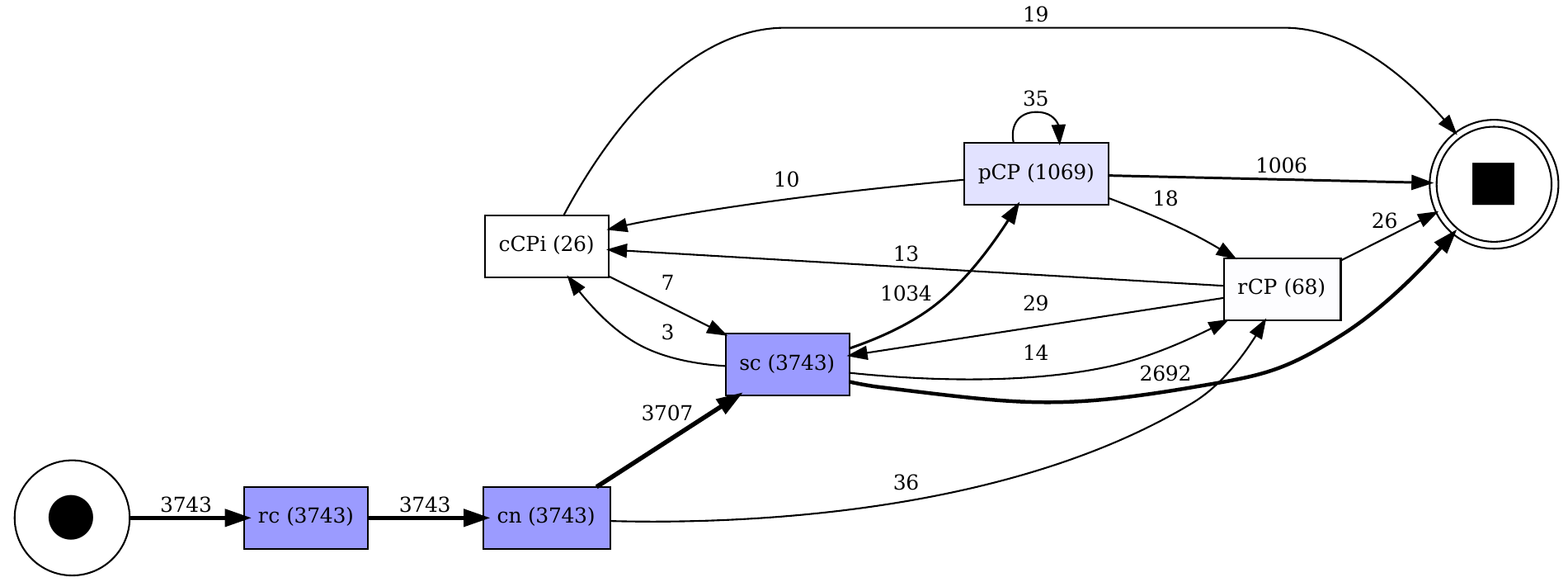}}
    \caption{The discovered Directly-Follows Graph (DFG) of claim part identification process, showing how AI scaled Claim Part identification.}
    \label{fig:overalldfg}
    \vspace{-1.5\baselineskip}	
\end{figure}

\figurename~\ref{fig:overalldfg} shows the DFG discovered by flattening the log on the claim object type without any filtering. 
In total, \csDFGrc{} claims have been registered (\textit{rc}) in this line of business over \caseDataCollectionPeriodInMonth{} months. 
Among these claims, claim handlers identified claim parts for only \csHDFGrCP{} claims, i.e., \textit{rCP}, while AI scanned all claims and identified \csDFGpCPIncomingFlow{} claim parts. Please note that the loop in \textit{pCP} occurred because a few prediction results were registered multiple times due to the use of message queues in the architecture, where messages timed out. Therefore, the incoming flow shows the correct number of unique predictions.

As can be seen, claim part investigators have created only a few cases for investigation due to the limited number of investigators currently handling these claim parts. 
The number of investigators was sufficient for the traditional process, as they did not investigate all identified claim parts but selected the most important ones based on certain claim features. 
However, AI significantly scaled the identification process, leading stakeholders to recognize \textit{the need to employ more investigators to handle the increased workload}. 
This demonstrates that while AI can enhance scalability, it can also shift bottlenecks within the process. 

These findings highlight the importance of assessing process reengineering efforts when implementing AI-driven improvements, as other parts of the process may be affected and require additional adjustments. 
\textit{This highlights the value of analyzing the end-to-end process when improving a part of the process using any AI-driven approach, as improving one specific point in the process may not add value to the organization}. 

The identified changes were considered highly valuable by stakeholders and the company, who plan to extend AI usage to other lines of business. 
Additional questions arose during the evaluation phase, and \textit{OCEL 2.0 proved beneficial by eliminating the need for repeated data extraction and integration for each analysis}, as it retains information about all object types. \textit{OCEL 2.0 also enabled us to query and filter different relations using qualifiers defined for object-to-object and event-to-object relations}, without which we needed to extract and integrate the data to an event log from scratch, increasing data processing time significantly for every analysis. 
The business was interested in investigating the process and answering questions (Q\ref{q:1}-Q\ref{q:4}) about how humans and AI contributed to claim part identification. We present the analysis for these questions below.

\vspace{0.5\baselineskip}	
\noindent\textbf{\textit{Q1) How have claim handlers effectively identified claim parts?}}

\begin{figure}[t!]
    \centering
    \adjustbox{center}{\includegraphics[width=1.2\linewidth]{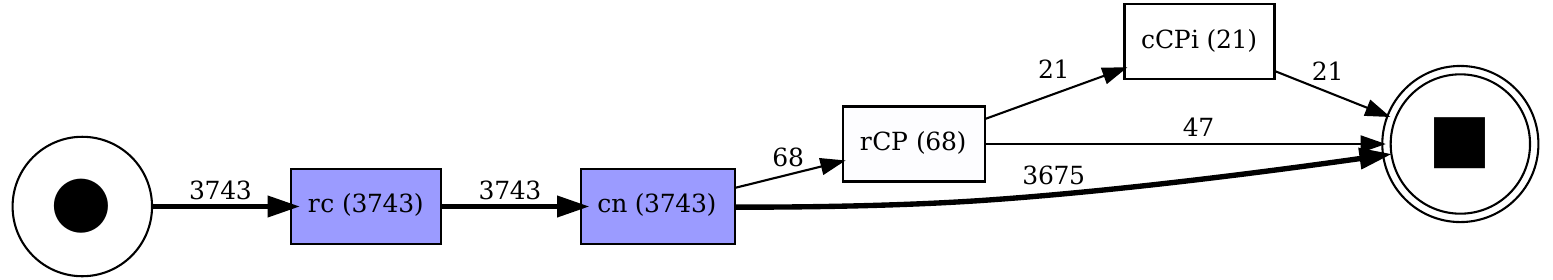}}
    \caption{The discovered DFG depicting human performance in identifying Claim Parts and investigating identified cases.}
    \label{fig:dfgHuman}
    \vspace{-1.5\baselineskip}	
\end{figure}

To answer this question, we filtered out ``scan claim" (\textit{sc}) and ``predict Claim Part" (\textit{pCP}) events from the log, which were only performed by AI.
Additionally, we retained only ``create Claim Part investigation" (\textit{cCPi}) events where its claim number could be found among reported Claim Parts (\textit{rCP}).
\figurename~\ref{fig:dfgHuman} presents the DFG obtained after filtering the OCEL accordingly and flattening the log based on the claim object type.
As can be seen, claim handlers identified \csHDFGrCP{} claim parts from \csDFGrc{} claims, indicating that \csHDFGrCPpercentage{} of claims contain claim parts. 
The business experts we interviewed confirmed this percentage, which is too low relative to the expected number of claim parts. This was a key motivation for initiating this project—to scale claim part identification.
From these identified claim parts, claim part investigators created \csHDFGcCPi{} cases for further investigation. 
It is worth mentioning that investigators do not create an investigation case for every identified claim part; instead, they select cases based on specific business criteria.

\vspace{0.5\baselineskip}	
\noindent\textbf{\textit{Q2) How has AI effectively identified claim parts?}}

\begin{figure}[b!]
    \centering
    \adjustbox{center}{\includegraphics[width=1.2\linewidth]{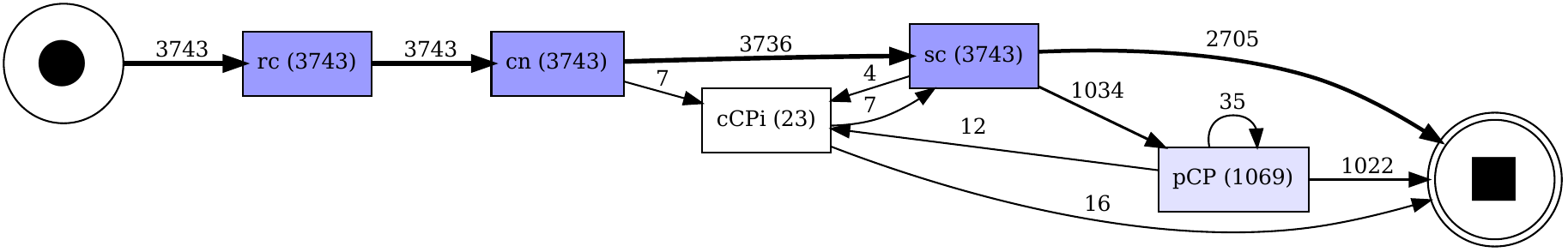}}
    \caption{The discovered DFG depicting AI performance in identifying Claim Parts.}
    \label{fig:dfgAI}
    % \vspace{-1\baselineskip}	
\end{figure}

To answer this question, we filtered out the ``report Claim Part" (\textit{rCP}) events from the log, which were performed solely by claim handlers. 
Additionally, we retained only ``create Claim Part investigation" (\textit{cCPi}) events where its claim number could be found among predicted Claim Parts (\textit{pCP}).
The reason for such filtering was that some claim part investigations may have been created based on claim handler reports before the AI would scan the claims. As a result, the AI would predict the claim part only after the investigation had already been created. Therefore, the \textit{cCPi} event would not be attributed to the AI, even though the AI would identify such cases.
\figurename~\ref{fig:dfgAI} presents the DFG obtained after filtering the OCEL accordingly and flattening the log based on the claim object type.

As can be seen, AI identified \csAIDFGpCPIncomingFlow{} claim parts (see the frequency of incoming flow to \textit{pCP}) from \csDFGrc{} claims, indicating that \csAIDFGrCPpercentage{} of claims contain claim parts. This percentage is expected by the business experts we interviewed, showing the success of claim part identification. Among these identified claim parts, the investigators created \csAIDFGcCPi{} cases for further investigation. 
Please note that the outgoing and incoming flows to \textit{sc} show the order of events that happened in different cases and do not indicate loops. 

It is evident from this figure that investigators created some claim parts before AI scanned the claims. Such a shift in their work does not affect the AI outcome, as AI did not know about the created items. Please note that as some \textit{cCPi} were not related to AI (as they were created before AI scanned the claim), they would have been missed if one filtered the traditional log by including only \textit{cCPi} events that are related to AI through an attribute. 
Thus, \textit{it was experienced that without having OCEL, we could either end up with incorrect analysis or need to perform a complex data processing and re-extracting the log to answer this question}. 

If we compare the AI-identified claim parts ratio with the human-identified claim parts ratio, we can conclude that the process has scaled \caseAIvsHumanCPIdentifiedDiffernece{}, a finding confirmed by the business when investigating the cases.
However, we do not observe a significant difference when examining the created claim part investigation (\textit{cCPi}).
Upon discussing this issue with business stakeholders, we realized that \textit{a lack of investigators is limiting the scalability of the claim part management process}. Since \textit{cCPi} initiates this process, its impact is constrained by this bottleneck.
This is an interesting finding, as it highlights how the successful adoption of AI in scaling the identification process creates a bottleneck in another process (i.e., the claim part management process). The company is now working to scale this aspect of the process to fully leverage the benefits of AI-identified claim parts by enabling the processing of a greater volume.

\vspace{1\baselineskip}	
\noindent\textbf{\textit{Q3) Among the claim parts under investigation, how many claim parts did AI fail to identify but were accurately identified by claim handlers?}}

\begin{figure}[b!]
    \centering
    \adjustbox{center}{\includegraphics[width=1.3\linewidth]{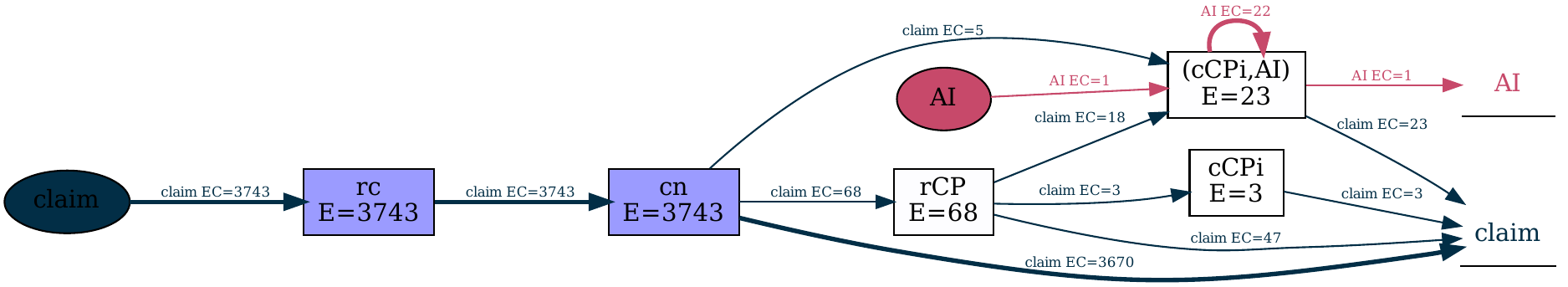}}
    \caption{The discovered OC-DFG separating cases identified by humans alone (\textit{cCPi E=3}) and those identified by AI or AI and humans (\textit{(cCPi,AI) E=23}).}
    \label{fig:dfgHumanNotAI}
    % \vspace{-1\baselineskip}	
\end{figure}

The claim parts that are under investigation are considered truly identified claim parts by the business as an investigator has approved them by creating an investigation request for them (\textit{cCPi}). Thus, the business was interested in investigating how the process looks like for  \csDFGcCPi{} claim parts for which the investigation request is made (\textit{cCPi} in \figurename~\ref{fig:overalldfg}).
To answer this question, we discovered an OC-DFG by applying some specific filtering, as explained below (see \figurename~\ref{fig:dfgHumanNotAI}).

The overall OC-DFG could answer the question as well, but stakeholders considered it very complex and impractical as it required a lot of effort to be interpreted. 
Thus, we excluded non-human activities from the log. 
Also, we only kept \textit{Claim} and \textit{AI} object types and unfolded~\cite{khayatbashi2024advancing,khayatbashi2024olap} the log based on \textit{cCPi} and \textit{AI}. 
The unfold operation enabled us to separate all investigation activities where AI was involved, helping users to visually separate the ones that were only performed by claim handlers. 
As can be seen from the OC-DFG, there were only \csAIMissed{} investigation cases where the claim handler reported the claim part, and AI missed them. 
The frequency of \textit{(cCPi,AI)} (23) represents the number of times that both AI or human and AI could identify claim parts.

\vspace{0.5\baselineskip}	
\noindent\textbf{\textit{Q4) Among the claim parts under investigation, how many claim parts did claim handlers fail to identify but were accurately identified by AI?}}

\begin{figure}[t!]
    \centering
   \adjustbox{center}{ \includegraphics[width=1.5\linewidth]{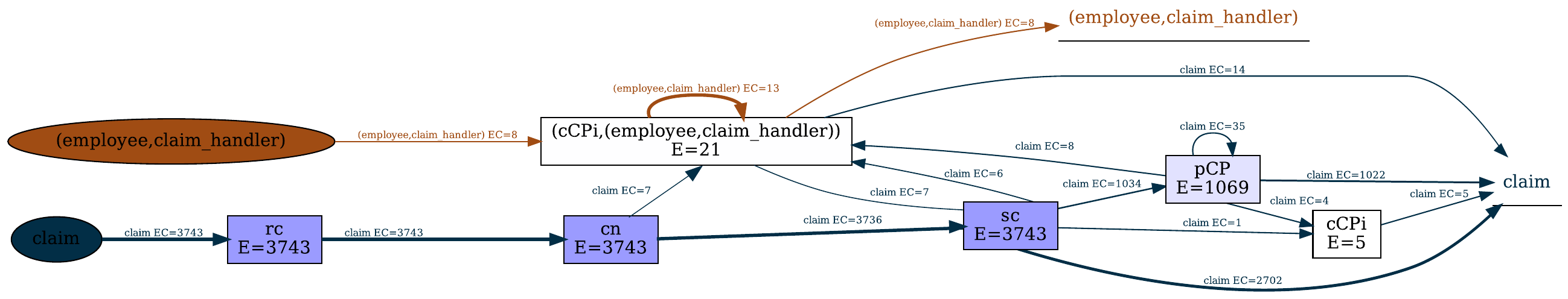}}
    \caption{The discovered OC-DFG separating cases identified by AI alone (\textit{cCPi E=5}) and those identified by humans or AI and human (\textit{(cCPi,(employee,claim\_handler)) E=21}).}
    \label{fig:dfgAINotHuman}
    \vspace{-1.5\baselineskip}	
\end{figure}

\begin{figure}[h!]
    \centering
    \begin{subfigure}[b]{0.3\textwidth}
        \centering
        \begin{tikzpicture}
            % Define colors
            \definecolor{lightpink}{RGB}{255,200,200}
            \definecolor{lightblue}{RGB}{200,200,255}

            % Draw and fill circles with transparency
            \fill[lightpink, opacity=0.5] (0,0) circle(1.5);  % AI Circle
            \fill[lightblue, opacity=0.5] (1.5,0) circle(1.5); % Claim Handler Circle
            
            % Draw outlines
            \draw[thick] (0,0) circle(1.5);  
            \draw[thick] (1.5,0) circle(1.5);  

            % Labels outside the circles
            \node at (-0.5, 1.7) {\textbf{AI}};  
            \node at (1.9, 1.7) {\textbf{Claim Handler}};  

            % Numbers inside the Venn diagram
            \node at (-0.7, 0) {\textbf{\csHMissed{}}};    % AI only
            \node at (2.2, 0) {\textbf{\csAIMissed{}}};     % Claim Handler only
            \node at (0.75, 0) {\textbf{\csAIandHuman{}}};   % Both AI and Claim Handler
        \end{tikzpicture}
        % \caption{}
    \end{subfigure}%
    \caption{Number of investigation created based on AI \& Human identification.}
    \label{fig:venn}
    \vspace{-2.5\baselineskip}	
\end{figure}
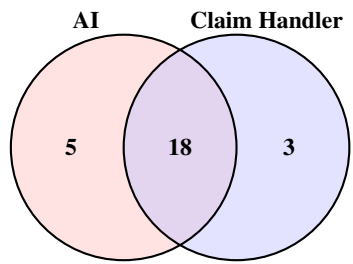

To answer this question, we only kept \textit{Claim} and \textit{(employee, claim\_handler)} object types and unfolded the log based on \textit{cCPi} and \textit{(employee, claim\_handler)}. We excluded \textit{rc} as it only performed by the claim handler, then discovered the OC-DFG (see \figurename~\ref{fig:dfgAINotHuman}).
As can be seen, AI identified \csHMissed{} cases that claim handlers missed among open to be investigated cases. 
\figurename~\ref{fig:venn} shows how many investigations were created based on identified cases by claim handlers and AI. 
As can be seen, 
AI and claim handlers identified \csAIandHuman{} cases among cases under investigation, where claim handlers missed identifying \csHMissed{} cases, and AI missed identifying \csAIMissed{} cases.
The result are all evaluated by discussing identified cases with stakeholders and verifying the correctness based on source systems.

\vspace{-1.2\baselineskip}	

\vspace{-0.1\baselineskip}	
\subsection{Lessons learned and discussion}
The integration of AI into business processes offers significant opportunities for efficiency gains but also introduces new challenges. We summarize our key findings below. 

\vspace{-0.7\baselineskip}	
\subsubsection{AI’s Impact on Business Process Dynamics}
AI-driven process reengineering does not guarantee efficiency improvements across the entire workflow. While AI can remove bottlenecks in one area, it may create new constraints elsewhere, necessitating process-wide adjustments:
\begin{itemize}[leftmargin=*]
    \item AI successfully eliminated bottlenecks in claim identification but created a backlog in claim investigation, requiring reallocation of resources and workforce adjustments.
    \item Scaling a single process step does not automatically lead to overall efficiency gains and adding business value. Organizations must ensure that downstream processes can handle the increased workload, either through automation, workforce expansion, or process redesign.
    \item AI-generated workload growth may exceed human capacity, highlighting the need for a holistic evaluation of AI adoption beyond individual task optimization.
\end{itemize}

\vspace{-0.7\baselineskip}	
\subsubsection{Challenges in Process Mining Visualization and Usability}
While process mining provides deep insights, effectively communicating these findings to business stakeholders remains a challenge. Several key visualization issues were identified:
\begin{itemize}[leftmargin=*]
    \item \textbf{Customization is crucial to usability.} Full object-centric models are too complex for non-technical users. Providing both simplified flat models and more detailed object-centric views improves adoption and decision-making.
    \item \textbf{Dynamic exploration capabilities are needed.} Drill-down and unfolding operations enable targeted analysis, reducing the need for repeated log reconstruction and enhancing efficiency.
    \item \textbf{Current open-source tools lack flexibility.} Existing open-source process mining tools do not offer the customization required for seamless stakeholder engagement. Improved visualization frameworks are necessary to bridge the gap between technical insights and business decision-making.
\end{itemize}

\vspace{-1\baselineskip}	
\section{Conclusion}\label{sec:conclusion}
\vspace{-0.5\baselineskip}	
This study demonstrates the application of AI-driven automation in scaling business processes within the insurance sector, specifically in the claims part identification process. By leveraging Large Language Models (LLMs) and Object-Centric Process Mining (OCPM), we systematically evaluated how AI enhances process scalability and efficiency. Our findings indicate that AI significantly increased the number of identified claim parts, reducing the manual workload of claim handlers. However, this success also introduced new bottlenecks, particularly in the claims investigation phase, highlighting the need for holistic process redesigns when integrating AI solutions.

The evaluation using OCPM provided valuable insights into the AI-driven transformation, allowing for detailed process analysis and performance tracking. However, our study also uncovered challenges in effectively communicating object-centric process models to stakeholders. While OCPM enabled in-depth analyses, traditional process representations were more understandable and acceptable for business users. These findings emphasize the need for improved visualization techniques to bridge the gap between complex process mining insights and practical business decision-making.

\bibliographystyle{plain} 
\bibliography{main} 

\end{document}